\documentclass[conference]{IEEEtran}

\usepackage{amsmath,amssymb,amsfonts}
\usepackage{algorithmic}
\usepackage{graphicx}
\usepackage{textcomp}
\usepackage{xcolor}
\usepackage{subcaption}
\usepackage{cite}
\usepackage{enumitem}
\def\BibTeX{{\rm B\kern-.05em{\sc i\kern-.025em b}\kern-.08em
    T\kern-.1667em\lower.7ex\hbox{E}\kern-.125emX}}
\begin{document}

\title{Towards Robust Autonomous Landing Systems: Iterative Solutions and Key Lessons Learned}


\author{\IEEEauthorblockN{1\textsuperscript{st} Sebastian Schroder}
\IEEEauthorblockA{\textit{School of Computing} \\
\textit{Macquarie University}\\
Sydney, Australia \\
sebastian.schroder@mq.edu.au}
\and
\IEEEauthorblockN{2\textsuperscript{nd} Yao Deng}
\IEEEauthorblockA{\textit{School of Computing} \\
\textit{Macquarie University}\\
Sydney, NSW, Australia \\
yao.deng@mq.edu.au}
\and
\IEEEauthorblockN{3\textsuperscript{rd} Alice James}
\IEEEauthorblockA{\textit{School of Computing} \\
\textit{Macquarie University}\\
Sydney, NSW, Australia \\
alice.james@mq.edu.au}
\and
\IEEEauthorblockN{4\textsuperscript{th} Avishkar Seth}
\IEEEauthorblockA{\textit{School of Computing} \\
\textit{Macquarie University}\\
Sydney, NSW, Australia \\
avishkar.seth@mq.edu.au}
\and
\IEEEauthorblockN{5\textsuperscript{th} Kye Morton}
\IEEEauthorblockA{\textit{Skyy Network} \\
\textit{Skyy Network} \\
Brisbane, QLD, Australia\\
kye@skyy.network}
\and
\IEEEauthorblockN{6\textsuperscript{th} Subhas Mukhopadhyay}
\IEEEauthorblockA{\textit{School of Computing} \\
\textit{Macquarie University}\\
Sydney, NSW, Australia \\
subhas.mukhopadhyay@mq.edu.au}
\and
\IEEEauthorblockN{7\textsuperscript{th} Richard Han}
\IEEEauthorblockA{\textit{School of Computing} \\
\textit{Macquarie University}\\
Sydney, NSW, Australia \\
richard.han@mq.edu.au}
\and
\IEEEauthorblockN{8\textsuperscript{th} Xi Zheng}
\IEEEauthorblockA{\textit{School of Computing} \\
\textit{Macquarie University}\\
Sydney, NSW, Australia \\
james.zheng@mq.edu.au}
}

\maketitle

\begin{abstract}

Uncrewed Aerial Vehicles (UAVs) have become a focal point of research, with both established companies and startups investing heavily in their development. This paper presents our iterative process in developing a robust autonomous marker-based landing system, highlighting the key challenges encountered and the solutions implemented. It reviews existing systems for autonomous landing processes, and through this aims to contribute to the community by sharing insights and challenges faced during development and testing.

\end{abstract}




\begin{IEEEkeywords}
Drone, UAV, Autonomous, Machine Learning, Landing
\end{IEEEkeywords}




 
\section{Introduction}
  
Autonomous landing of Uncrewed Aerial Vehicles (UAVs) represents a critical and core aspect for developing the reliability and safety of UAV operations and paves the way for more complex and ambitious applications of drone technology in both civilian and military domains. Applications such as package delivery services \cite{zieher2024parceldelivery}  and infrastructure inspections \cite{schofield2020inspection}  benefit from improved landing systems. Autonomous landing systems can be broadly categorised into two types: marker-based \cite{mohammad2017arucolanding} and marker-less \cite{krpec2024markerless} . Marker-based systems utilise fiducial markers or beacons at the landing site for precise positioning, while marker-less systems aim to identify suitable landing areas without pre-placed markers. This industry experience paper focuses on our development, and testing of marker-based autonomous landing systems. 
\begin{figure}
    \centering
    \includegraphics[width=0.5
    \textwidth]{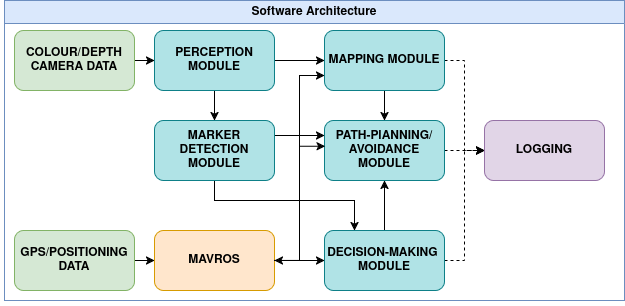}
    \caption{Software Architecture of Autonomous Landing System}
    \label{fig:system_architecture}
\end{figure} 

We propose a multi-module architecture for autonomous landing, enabling aircraft to safely and precicely land without direct human interference. Figure~\ref{fig:system_architecture} shows the overall system architecture of the drone system, with this paper predominantly focusing on the marker detection and local path planning software modules. These modules are dynamically loaded on the onboard companion computer.
The goal of this paper as is to address key challenges in marker detection accuracy across diverse environments and weather conditions, as well as show improvements in collision-free path planning during both the marker search and landing phases.


In this paper, we make several key contributions. First, we demonstrate that our neural-network based marker detection significantly out-performs conventional OpenCV \cite{opencv_library} implementations in terms of accuracy and robustness. Second, we show that our path planning approach based on RRT* from OMPL \cite{Sucan2012TheOM} achieves substantial improvements over traditional A* \cite{hart1968astar} algorithms, particularly in complex environments. Finally, we provide a critical analysis of the current limitations of our system, identifying promising directions for future research. Our experimental results validate these contributions across a range of realistic scenarios in Software-in-the-Loop (SIL), Hardware-in-the-Loop (HIL) testing, and real-world testing.


\section{Related Work}
\begin{figure*}[t]
    \centering
    \includegraphics[width=1.0\linewidth]{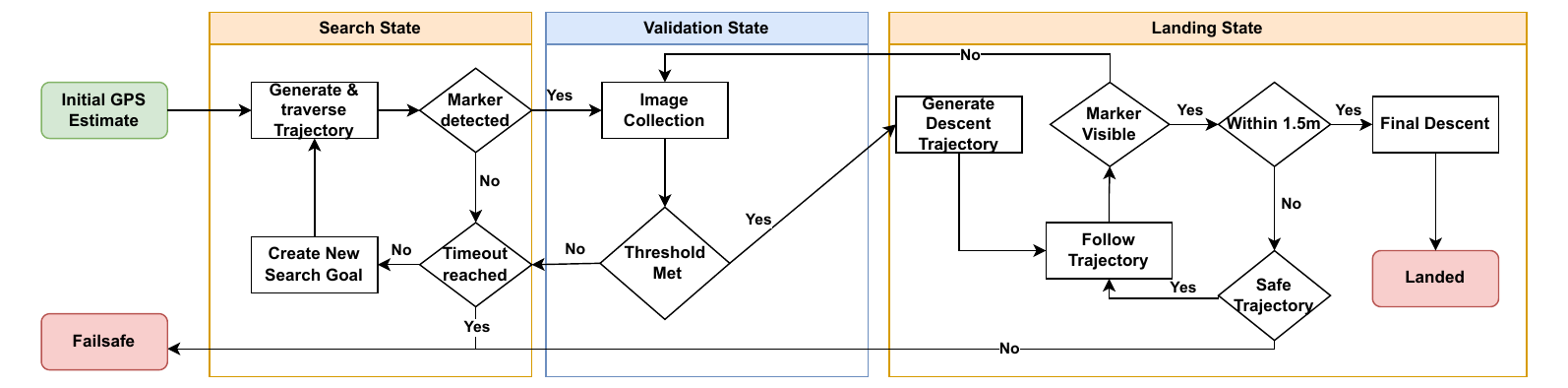}
    \caption{Decision Making Module State Machine}
    \label{fig:state-machine}
\end{figure*}
\label{sec:related_work}
\subsection{Marker Detection}
Safe autonomous UAV landing is crucial for applications like delivery, surveying, and return-to-home operations. According to \cite{xin2022vision}, landing can be categorised into three types: static targets, moving targets, and complex environments without predefined targets. This paper focuses on landing on static targets, commonly used in delivery applications.

Traditional autonomous landing approaches relied on "H" \cite{saripalli2002vision}, "T" \cite{tsai2006terminal}, or circular patterns \cite{lange2009vision}, which feature simple visual characteristics for easy recognition. with position and orientation computed using identifiable features. Recently, Augmented Reality Markers (ArUco markers)\cite{garrido2014automatic} have gained popularity in UAV landings, offering higher accuracy, lower rotational ambiguity, and the ability to encode IDs with minimal data. However, adverse environmental conditions (fog, sun glare, or shadows~\cite{verma2024comparative}) and low marker resolution can decrease detection performance.  Deep learning-based detection approaches have emerged as a more robust alternative for marker detection, such as Mask R-CNN~\cite{blachut2022automotive}, and a variety of YOLO versions~\cite{li2020aruco, zhao2023vision}. Deep neural networks can learn complex visual patterns and can better handle traditionally challenging conditions such as poor lighting, motion blur, colour aberration or partial obstruction.

\subsection{Path Planning}
The path planning module generates collision-free trajectories to target locations. Initially, we employed EGO-Planner \cite{Zhou2020EGOPlannerAE}, which utilises the A* algorithm for path planning. While this approach proved effective in many scenarios, our testing revealed two significant limitations:
\begin{itemize}[topsep=0pt, itemsep=-2pt, parsep=0pt, partopsep=0pt, leftmargin=*, labelindent=0pt, labelsep=0.5em]
    \item When encountering large obstacles, such as buildings, the A* algorithm often failed to find viable solutions within the constraints of the search pool size and real-time response requirements. This limitation became particularly apparent in urban environments where the UAV needed to navigate around substantial structures.
    \item EGO-Planner relies on local obstacle information for path planning, which occasionally led to problematic trajectories. For instance, when operating above wooded areas, the planner would create an optimal path that went through at-the-time unseen obstacles and could then become trapped within the foliage of a tree.

\end{itemize}
To mitigate these limitations, we developed a new path planner based on the Open Motion Planning Library (OMPL) \cite{Sucan2012TheOM}, which provides multiple sample-based path planning algorithms. Specifically, we adopted the RRT* (Rapidly-exploring Random Tree Star) algorithm for path planning. By integrating the global obstacle information stored in our octree-based occupancy map, the path planner can generate more robust trajectories that account for the complete environmental structure, reducing the likelihood of problematic paths.

\section{Methodology}
\label{sec:method}

\begin{figure*}
    \centering
    \includegraphics[width=1.0\textwidth]{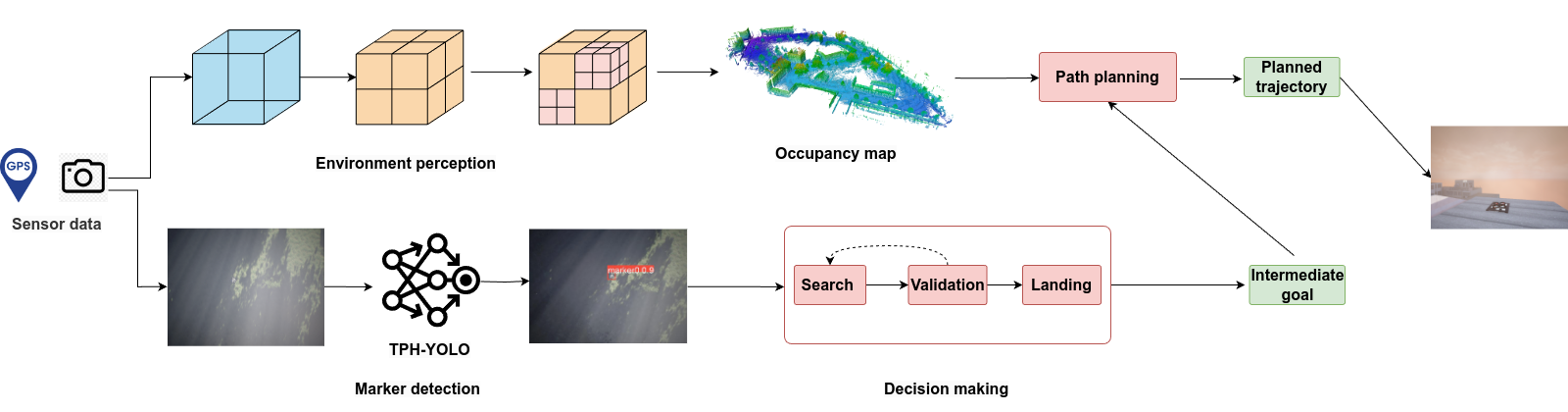}
    \caption{Workflow of the marker-based autonomous landing system}
    \label{fig:workflow}
\end{figure*}

Figure~\ref{fig:workflow} illustrates the overall methodology and workflow of our proposed marker-based autonomous landing system. The system's environment perception module and mapping module uses depth sensor data to generate an octree-based occupancy map to enable collision avoidance,  while scanning below for an ArUco marker using a color camera. Based on these data, the decision-making module determines the appropriate action. This may include continuing the search for the marker, validating a detected marker, or initiating the landing process. Finally, whenever movement is required, the path planning module generates a collision-free trajectory to this intermediate goal, taking into account the obstacle information stored in the occupancy map. 

\subsection{Marker Detection Module}
Our marker detection module originally used a traditional fixed algorithm detection method using the OpenCV library \cite{opencv_library} to detect fiducial ArUco markers. However, early testing revealed limitations in accurately detecting and identifying individual ArUco markers, particularly during high altitude flight, with partial marker occlusion (due to shadows or other environmental factors), and under challenging lighting conditions such as sun glare that affected image quality.

\begin{figure*}
    \centering
    \includegraphics[width=1\linewidth]{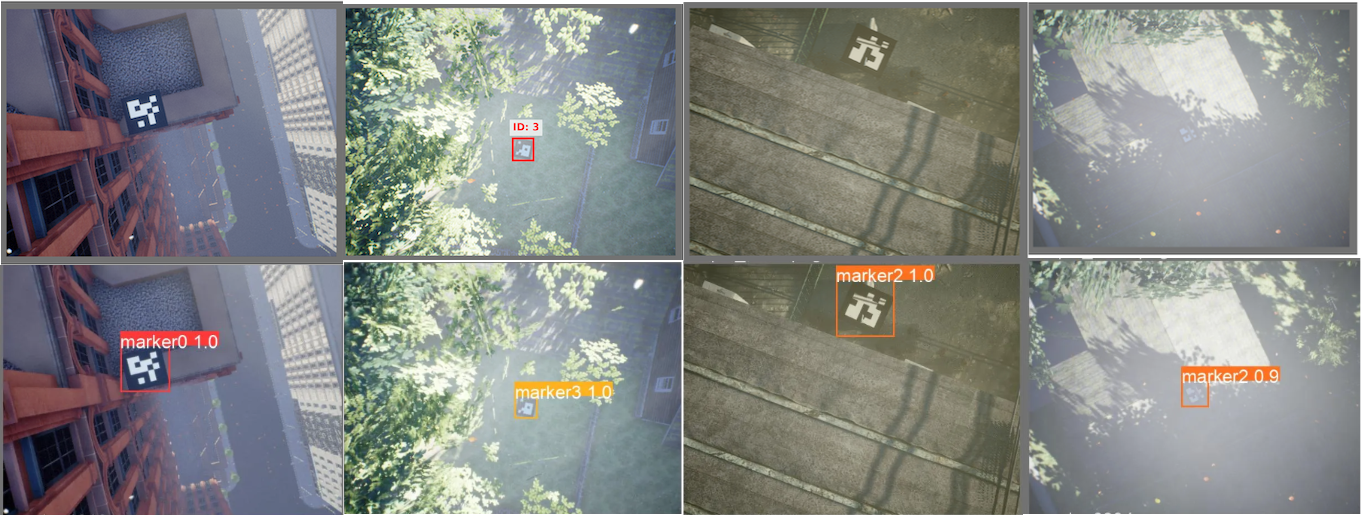}
    \caption{Examples of marker detection results. The top row is from OpenCV and the bottom row is from TPH-YOLO}
    \label{fig:marker_detection}
\end{figure*}

To address these limitations and enhance the effectiveness of our marker detection system, we transitioned to using a deep learning model,  adopting TPH-YOLO \cite{Zhu2021TPHYOLOv5IY}, a variant of YOLO-V5 that incorporates transformer blocks to optimise the detection of small objects. To develop a robust marker detection model, a dataset of ArUco markers were created using data from five customised maps in the AirSim simulator \cite{shah2018airsim}, a subset of those maps can been seen in figure \ref{fig:marker_detection}. AirSim was chosed for its ability to render high-fidelity environments, surpassing other simulators like Gazebo in visual quality. Moreover, as AirSim is implemented as an Unreal Engine plugin, we could easily configure diverse simulation maps based on Unreal game projects, encompassing various environments such as rural and urban areas.

Our dataset consists of images from these simulated environments, in which markers were placed in unique positions and orientations, various weather conditions were simulated, and the drone operated at various orientations and heights. To further enhance the model, we applied image filters, including random adjustments to brightness and contrast, as well as the addition of Gaussian noise. This combination significantly improved our marker detection capabilities, enabling more reliable autonomous landing across a wide range of real-world conditions.
 
\subsection{Mapping Module}


The Mapping module constructs and maintains a representation of the surrounding environment using processed point cloud data. This data is stored in an occupancy map, in which 3D space is discretised into voxels (volumetric pixels), where each voxel is a boolean value indicating whether the space it represents is occupied by an obstacle or not. 

Initially, this occupancy map was implemented using a three-dimensional static grid array structure, akin to EGO-Planner~ \cite{Zhou2020EGOPlannerAE}. This implementation offers fast access to occupancy information, however granularity and effective memory usage were mutually exclusive. 

To address this, we transitioned to an octree-based representation using OctoMap \cite{Hornung2013OctoMapAE}, which hierarchically partitions the 3D space, allowing for efficient compression of large empty or occupied areas into single nodes. This approach significantly reduces memory usage while maintaining the ability to represent fine details where needed. Additionally, OctoMap's probabilistic update mechanism naturally handles sensor noise and uncertainty, making it well-suited for real-world applications.

\subsection{Path Planning Module}
The path planning module generates collision-free trajectories to targets. We initially used EGO-Planner \cite{Zhou2020EGOPlannerAE} with its A* algorithm, but testing revealed two key limitations:
\begin{itemize}[topsep=0pt, itemsep=-2pt, parsep=0pt, partopsep=0pt, leftmargin=*, labelindent=0pt, labelsep=0.5em]
    \item Large obstacles often caused the A* algorithm to fail in finding viable paths within search constraints and real-time requirements, particularly in urban environments.
    \item Reliance on local obstacle information sometimes produced problematic trajectories, such as paths through temporarily unseen obstacles that could trap the UAV in tree foliage.
\end{itemize}
To address these issues, we implemented a new planner using the Open Motion Planning Library (OMPL) \cite{Sucan2012TheOM}, specifically the RRT* algorithm. By incorporating global obstacle data from our octree-based map, this approach generates more robust trajectories that account for complete environmental structure.

\subsection{Decision-making Module}
The decision-making module manages the UAV's behaviour by transitioning between distinct operational states: search, validation, and landing, which can be seen in Figure \ref{fig:state-machine}.

\textbf{Searching State}
The UAV explores the target area to locate the landing marker. Upon reaching this position, if no marker is detected, the drone attempts a spiral search pattern. If a marker is not found, the system enters a fail-safe. If a marker is found, the system enters the validation stage.

\textbf{Validation State} 
Once a theoretical marker is found, the UAV will hover and collect a series of detection results across multiple frames; if a threshold is met, validation is successful and it enters landing state, otherwise it returns to the searching state.

\textbf{Landing State}
In the landing state, the module generates below a sequence of descending waypoints that guide the UAV to a safe landing at the validated marker. The goal of this sequence is to ensure the continuous visual contact with the marker and ensure adequate clearance from any detected obstacles. If any of the safety features mentioned above are breached, the system enters a failsafe state. This state involves aborting the current landing attempt and executing a pre-configured failsafe procedure, such as returning home or re-initiating the marker search (returning to the validation or search state as appropriate).

\textbf{Safety and Availability Considerations}
Designing the autonomous landing system involves balancing operational safety against successful landing availability. Stricter safety parameters like larger obstacle clearances, more rigorous marker validation, or sensitive failsafe triggers—reduce collision risks but can also decrease availability by causing more frequent landing aborts in challenging conditions (e.g., poor visibility). Conversely, relaxed constraints might increase landing success rates in difficult scenarios but elevate risk. Our system, particularly through its global path planning (RRT* with OctoMap), multi-stage validation, and failsafe mechanism, prioritizes safety by aborting potentially unsafe attempts rather than risking failure, thus favouring safety over immediate task completion.

\section{Experiments}
\label{sec:experiment}
\subsection{Research Questions}
To evaluate the effectiveness and robustness of our proposed autonomous landing system, we designed a comprehensive set of experiments addressing the following research questions (RQs):

\begin{itemize}
    \item RQ1: How does the performance of our multi-module autonomous landing system compare to existing methods in Software-in-the-Loop (SIL) simulation testing?

    \item RQ2: To what extent does the system maintain its performance when running on real hardware in Hardware-in-the-Loop (HIL) testing?

    \item RQ3: What is the system's performance and reliability when deployed in real-world scenarios?
\end{itemize}

These research questions are designed to assess our system's capabilities across different levels of abstraction, from controlled simulations to practical, real-world applications.

\subsection{Benchmark}

\subsubsection{Simulation Environments}
We created 10 simulation maps using AirSim~\cite{shah2018airsim}, a high-fidelity drone simulator based on Unreal Engine (UE)~\cite{unrealengine}, encompassing both rural, suburban and urban areas.  For each map, we generated 10 distinct test scenarios, equally divided between normal and adverse weather conditions. This comprehensive set of 100 scenarios provides a robust framework for evaluating our autonomous landing system across a wide range of environmental conditions and challenges.

    \label{fig:gui}

\subsubsection{Autonomous Landing Systems}
We evaluated three generations of marker-based autonomous landing systems. The first generation system (MLS-V1) utilises an OpenCV-based marker detector without object avoidance capabilities. The second generation system (MLS-V2) incorporates TPH-YOLO~\cite{kalaitzakis2020experimental} for marker detection and Ego-planner~\cite{Zhou2020EGOPlannerAE} for generating collision-free paths. The third generation system (MLS-V3) combines TPH-YOLO for marker detection with Octomap~\cite{Hornung2013OctoMapAE} mapping and the RRT*~\cite{Karaman2011SamplingbasedAF} path planning algorithm.

\subsection{Experiment Settings}
\subsubsection{RQ 1}
For software-in-the-loop (SIL) testing, both the autonomous landing system and the simulation environment were executed on a single desktop computer. The UAV firmware (PX4) was simulated alongside the environment and software modules.
In each scenario, the drone was initialized at the map origin $(0, 0, 0)$, took off, and navigated towards a designated GPS target. The target marker, along with false positive markers, was placed within a defined radius of the target. Upon detecting the target marker, the drone attempted to land, completing the scenario.

We evaluated three versions of the autonomous landing system (MLS-V1, MLS-V2, and MLS-V3) over 100 generated scenarios. Each scenario was repeated three times to ensure result consistency. For every run, we recorded: (1) the deviation between the detected and actual marker positions, (2) the deviation between the landing and target marker positions, and (3) the number of collisions. The average of these metrics was computed to comprehensively assess each system's performance.

\begin{figure}
    \centering
    \begin{subfigure}[b]{0.48\textwidth}   
        \centering 
        \includegraphics[width=\textwidth]{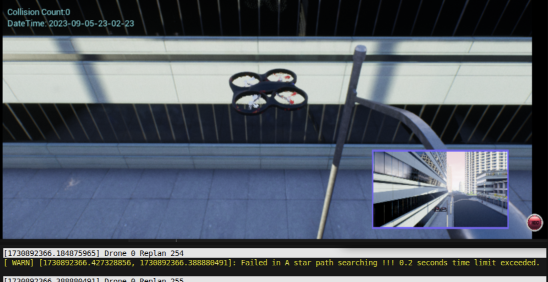}
        \caption{Path planning failure of MLS-V2 due to large obstacles}    
        \label{fig:ego_v2_bug}
    \end{subfigure}
    \hfill
    \begin{subfigure}[b]{0.48\textwidth}   
        \centering 
        \includegraphics[width=\textwidth]{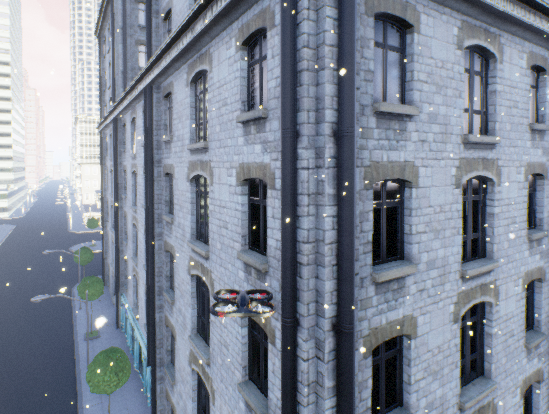}
        \caption{Collision during turning action while close to obstacle}    
        \label{fig:ego-v2-collision}
    \end{subfigure}
    
    \vspace{0.5cm}
    
    \begin{subfigure}[b]{0.48\textwidth}
        \centering
        \includegraphics[width=\textwidth]{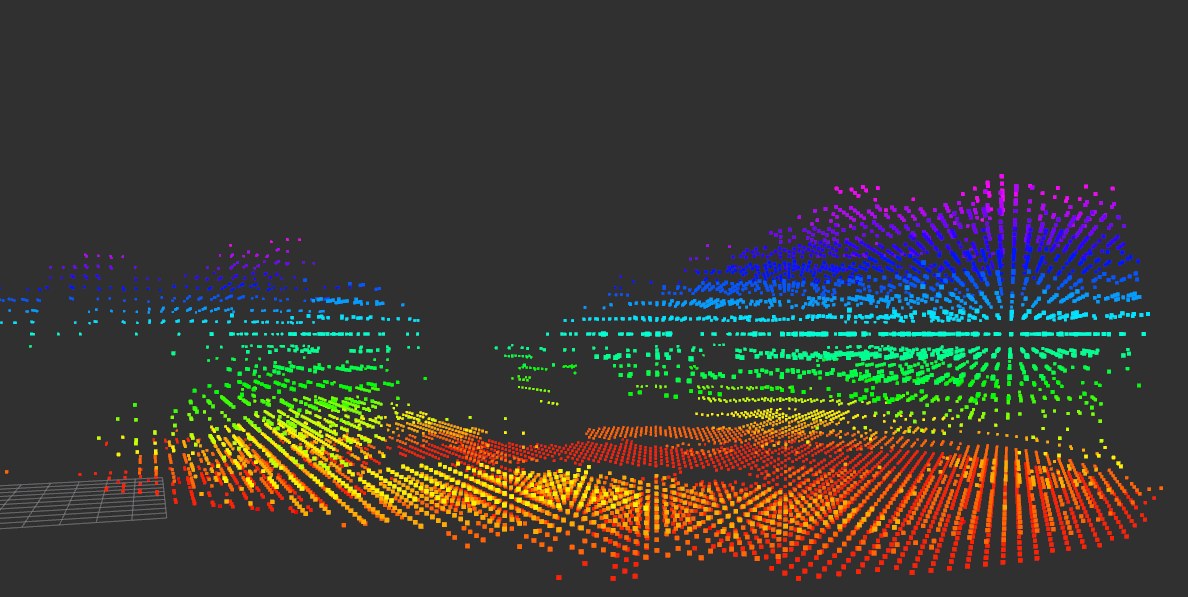}
        \caption{Erroneous pointclouds during IRL testing}    
        \label{fig:erroneous-pointclouds}
    \end{subfigure}
    \hfill
    \begin{subfigure}[b]{0.48\textwidth}  
        \centering 
        \includegraphics[width=\textwidth]{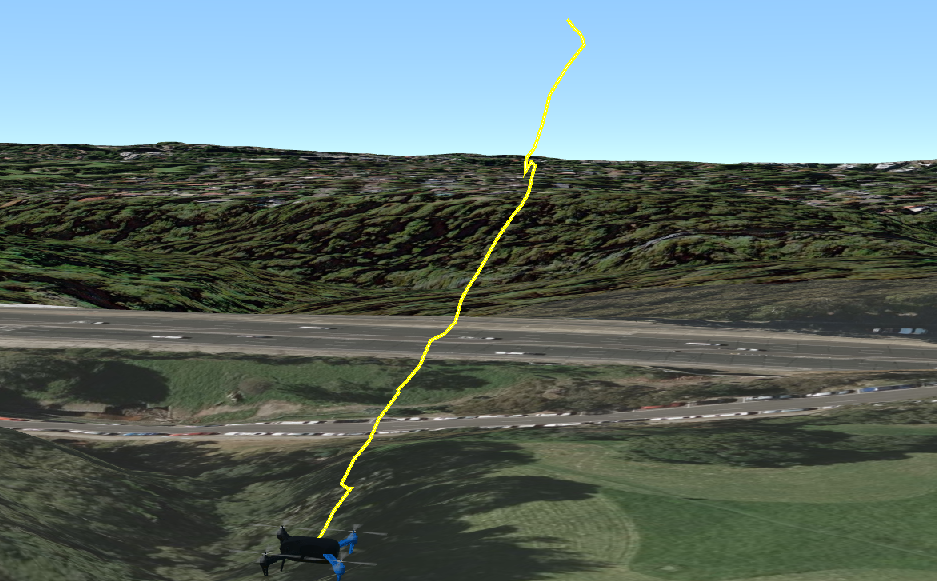}
        \caption{GPS drift during poor weather}    
        \label{fig:gps-drift-plot}
    \end{subfigure}
    
    \caption{Various issues experienced during testing} 
    \label{fig:issues-experienced-during-testing}
\end{figure}
    
\subsubsection{RQ 2}
For hardware-in-the-loop (HIL) testing, we adopted a distributed setup where the simulation environment and PX4 firmware were executed on a high-performance computer, while the autonomous landing system modules were deployed on an NVIDIA Jetson Nano~\cite{jetsonnano}, a compact edge device with a GPU and 4\,GB of memory. To maximize computational throughput and prioritize the performance of the autonomous landing algorithms, the Jetson Nano was configured to operate in its highest performance power mode (MAXN). To ensure the TPH-YOLO model could be processed efficiently on the edge device under this configuration, we optimized and converted it to the TensorRT format~\cite{tensorrt}, which significantly accelerates inference on NVIDIA GPUs. The testing procedure followed the same steps as in the SIL tests. In addition to the SIL performance metrics, we monitored the Jetson Nano's CPU/GPU usage and memory utilization to evaluate system efficiency under these resource-constrained, high-performance conditions.

\subsubsection{RQ 3}
For real-world testing, we used standard COTS components to rapidly develop a sensor-equipped drone. The platform consisted of an F450 quadcopter frame~\cite{F450} with a Jetson Nano (4\,GB) as the onboard companion computer. Initially equipped with a Pixhawk 2.4.8~\cite{pixhawk} flight controller, it was later replaced with a Cuav X7+ Pro for improved performance and enhanced sensor capabilities. A forward-facing Realsense D435~\cite{Realsense_D435} and a downward-facing D435i enabled marker and object detection, while a NEO 3 GPS~\cite{neo3} provided positioning data. Altitude was measured using a TFMini Plus LiDAR range sensor~\cite{TFMini-Plus}.

Each sensor was individually calibrated against known references. Subsystems were validated prior to full system testing, including manual flights in both GPS-enabled and GPS-denied environments and obstacle avoidance tests. During these experiments, system performance and resource usage on both the Jetson Nano and flight controller were recorded to assess real-world operation.

\section{Results}
\label{sec:results}
\subsection{SIL Results}

\begin{table}[!htpb]
\centering

\begin{tabular}{llll}
\hline
\textbf{\begin{tabular}[c]{@{}l@{}}Landing\\ System\end{tabular}} & \textbf{\begin{tabular}[c]{@{}l@{}}Successful \\ Landing Rate\end{tabular}} & \textbf{\begin{tabular}[c]{@{}l@{}}Failure rate due \\ to Collision\end{tabular}} & \textbf{\begin{tabular}[c]{@{}l@{}}Failure rate due\\ to poor landing\end{tabular}} \\ \hline
\textbf{MLS-V1}                                                   & 24.67\%                                                                     & 71.33\%                                                                           & \textbf{4.00\%}                                                                     \\
\textbf{MLS-V2}& 42.00\%                                                                     & 48.67\%                                                                           & 9.34\%                                                                              \\
\textbf{MLS-V3}& \textbf{84.00\%}                                                            & \textbf{3.33\%}                                                                   & 12.67\%                                                                             \\ \hline
\end{tabular}

\caption{Experiment Results of SIL Testing} 
\label{tab:sil-testing}
\end{table}
Table~\ref{tab:sil-testing} presents the experimental results from SIL testing across 150 scenarios. It's  sectioned based on the averaged results of each of the 3 system versions,  MLS-V1 had the highest overall failure rate, with 75.33\% of flights being considered a failure due mainly to collisions, which can be attributed to the poor to non-existant ability of the avoidance system. MLS-V2 improves upon V1 with 31\% lower overall collisions, however also experiences a small increase in failures during landing, which is attributed to the swap from OpenCV to YOLO based marker detection. Finally MLS-V3 shows the best overall performance, with a significant improvement in the overall landing rate. 
Among the failure causes,  catastrophic collisions are way down.  However due to the new avoidance system there's a slight uptick in the number of failures experienced during landing, which are recoverable.

Table~\ref{tab:marker-detection-results} shows a gradual improvement in marker detection performance. The deep learning methods in MLS-V2 and V3 significantly outperformed the OpenCV-based approach in MLS-V1. However, these models were not trained for marker orientation estimation, limiting their applicability. Notably, later models performed much better under challenging conditions, such as poor weather and obfuscation.
\begin{table}[!htpb]
\centering
\begin{tabular}{lll}
\hline
\textbf{\begin{tabular}[c]{@{}l@{}}Marker Detection\\ Results\end{tabular}} & \textbf{Implementation} & \textbf{\begin{tabular}[c]{@{}l@{}}False Negative\\ Rate (\%)\end{tabular}} \\ \hline
\textbf{MLS-V1}                                                             & OpenCV                  & 4.00\%                                                                      \\
\textbf{MLS-V2}                                                             & TPH-YOLO                & 2.67\%                                                                      \\
\textbf{MLS-V3}                                                             & TPH-YOLO                & \textbf{2.00\%}                                                             \\ \hline
\end{tabular}
\

\caption{Marker Detection Results}
\label{tab:marker-detection-results}
\end{table}

\begin{figure}[!htpb]
    \centering
    \includegraphics[width=1\linewidth]{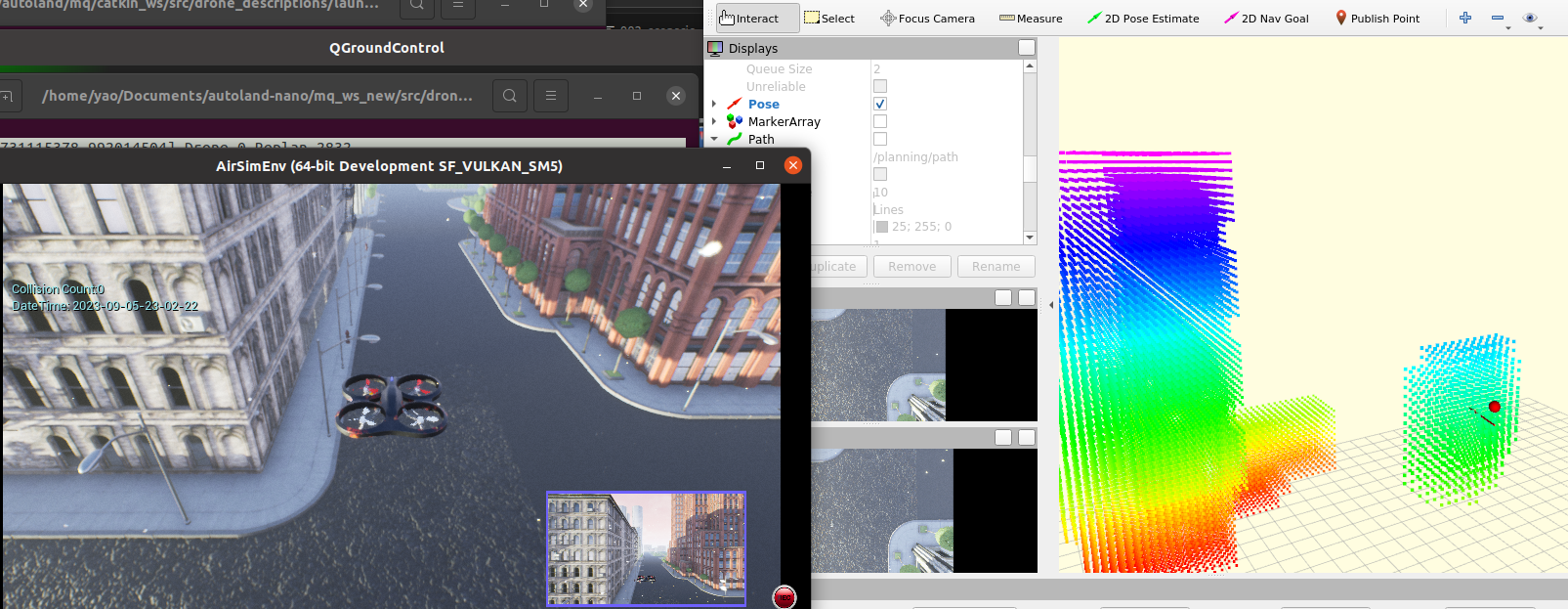}
    \caption{Inflated bounding box}
    \label{fig:sil-mapping}
\end{figure}

Failure patterns differed between versions. MLS-V2 collisions occurred primarily near buildings where objects were 'swallowed' by the bounding box (Figures \ref{fig:ego_v2_bug} \& \ref{fig:sil-mapping}), either invalidating all paths during safety checks or defaulting to unsafe straight-line paths. MLS-V3 failures happened mainly at sharp RRT* corners due to trajectory-following limitations, and when dynamic replanning occurred too late as the drone had already entered inflated obstacle boundaries (Figure \ref{fig:sil-mapping}).

\subsection{HIL Results}
Table~\ref{tab:hil-testing} shows the results from HIL testing across 150 scenarios for MLS-V3. 
\begin{table}[!htpb]

\begin{tabular}{llll}
\hline
\textbf{\begin{tabular}[c]{@{}l@{}}Landing\\ System\end{tabular}} & \textbf{\begin{tabular}[c]{@{}l@{}}Successful \\ Landing Rate\end{tabular}} & \textbf{\begin{tabular}[c]{@{}l@{}}Failure rate due \\ to Collision\end{tabular}} & \textbf{\begin{tabular}[c]{@{}l@{}}Failure rate due\\ to poor landing\end{tabular}} \\ \hline
\textbf{MLS-V3}                                                   & 72.00\%                                                                     & 14.00\%                                                                           & \textbf{6.00\%}                                                                     \\ \hline
\end{tabular}

\caption{Experiment Results of HIL testing}
\label{tab:hil-testing}
\end{table}

The success rate decreases predominantly from the increased number of collision failures.  This can be attributed to computational resource constraints on the Jetson Nano.  In these incidents, trajectories failed to create in time when the drone was heading towards a newly discovered obstacle, leading to increased collision rates. The HIL system consumes approximately 2.2GB out of the available 2.9GB memory, with all four CPU cores heavily utilised. The resource monitoring indicates that CPU processing power is the primary bottleneck for running the autonomous landing systems with multiple concurrent tasks.

\subsection{Real World Results}
For real-world testing, scenarios were simplified to fit within the limited airspace available. This subset of scenarios was designed to evaluate the effectiveness of the MLS-V3 model. Due to safety concerns, MLS-V1 and MLS-V2 were not tested, as their behavior near trees and water posed risks. 

\begin{figure}[!htpb]
    \centering
    \includegraphics[width=1\linewidth]{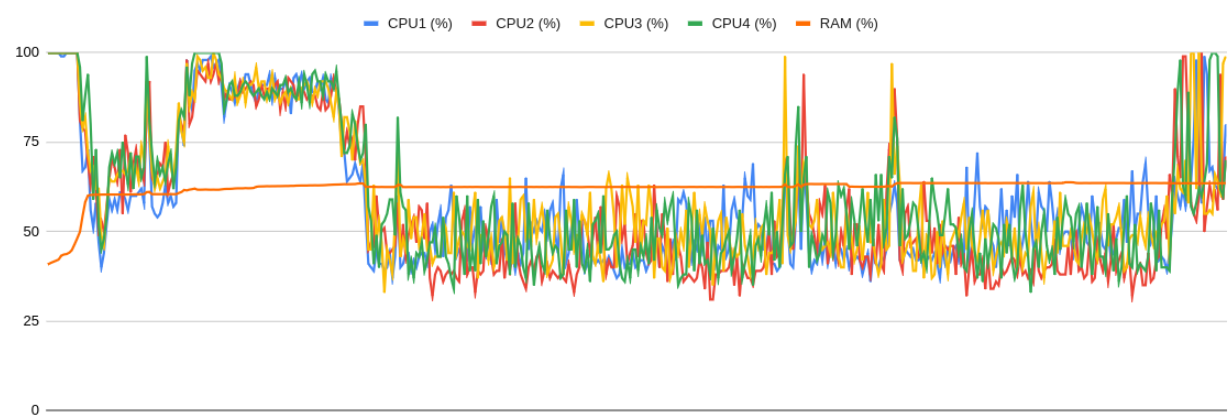}
    \caption{Real world Jetson Nano performance}
    \label{fig:rl-performance}
\end{figure}
Several new issues were observed. The most significant was GPS positioning drift (see figure \ref{fig:gps-drift-plot}), despite VDOP/HDOP values being within 2–8. This drift, likely caused by poor weather, affected the EKF state estimates, and caused erroneous points in the mapping module (see figure \ref{fig:erroneous-pointclouds})  . This could be mitigated by introducing an RTK base station, integrating a visual positioning system (though preliminary tests were unsatisfactory), or switching to an Off-board flight mode to allow relative positioning during critical phases such as the final descent. GPS drift also degraded mapping accuracy, occasionally preventing valid path generation.


Another issue was poor local positioning due to low-quality acceleration and rotational data, which was addressed by upgrading from Pixhawk 2.4.8 to the Cuav X7+ flight controller, featuring triple IMUs, dual barometers, and improved sensors.
RAM and CPU core resources were used noticably more than in HIL testing, which can be accounted for by the real time processing and communication of camera feeds (see Figure \ref{fig:rl-performance}).
Despite these challenges, the drone was able to land within 60\,cm of the marker on average, higher than the 25\,cm observed in SIL and HIL tests, primarily due to GPS inaccuracies and wind during the final descent.

\section{Conclusion}
\label{sec:conclusion}
We present the iterative development and evaluation of a multi-module UAV landing system, tested extensively through simulation, hardware-in-the-loop, and real-world experiments. This paper documents critical challenges and engineering solutions, offering actionable insights for advancing autonomous UAV landing systems. Our current modularization of software components lays the groundwork for future compositional verification and the adoption of the neurosymbolic paradigm~\cite{zheng2025neurostrata} for formal verification.

\clearpage


\bibliographystyle{IEEEtran}
\bibliography{references}

\begin{thebibliography}{10}
\providecommand{\url}[1]{#1}
\csname url@samestyle\endcsname
\providecommand{\newblock}{\relax}
\providecommand{\bibinfo}[2]{#2}
\providecommand{\BIBentrySTDinterwordspacing}{\spaceskip=0pt\relax}
\providecommand{\BIBentryALTinterwordstretchfactor}{4}
\providecommand{\BIBentryALTinterwordspacing}{\spaceskip=\fontdimen2\font plus
\BIBentryALTinterwordstretchfactor\fontdimen3\font minus \fontdimen4\font\relax}
\providecommand{\BIBforeignlanguage}[2]{{%
\expandafter\ifx\csname l@#1\endcsname\relax
\typeout{** WARNING: IEEEtran.bst: No hyphenation pattern has been}%
\typeout{** loaded for the language `#1'. Using the pattern for}%
\typeout{** the default language instead.}%
\else
\language=\csname l@#1\endcsname
\fi
#2}}
\providecommand{\BIBdecl}{\relax}
\BIBdecl

\bibitem{zieher2024parceldelivery}
E.~O. Simon~Zieher \emph{et~al.}, ``Drones for automated parcel delivery: Use case identification and derivation of technical requirements,'' \emph{Transportation Research Interdisciplinary Perspectives}, vol.~28, 2024.

\bibitem{schofield2020inspection}
O.~B. Schofield, K.~H. Lorenzen, and E.~Ebeid, ``Cloud to cable: A drone framework for autonomous power line inspection,'' in \emph{2020 23rd Euromicro Conference on Digital System Design (DSD)}, vol.~23, 2020, pp. 503--509.

\bibitem{mohammad2017arucolanding}
M.~F. Sani and G.~Karimian, ``Automatic navigation and landing of an indoor ar. drone quadrotor using aruco marker and inertial sensors,'' in \emph{2017 International Conference on Computer and Drone Applications (IConDA)}, 2017, pp. 102--107.

\bibitem{krpec2024markerless}
N.~S. Krypec~Blake, Valasek~John, ``Vision-based marker-less landing of an unmanned aerial system on moving ground vehicle,'' \emph{Journal of Aerospace Information Systems}, vol.~21, 2024.

\bibitem{opencv_library}
G.~Bradski, ``{The OpenCV Library},'' \emph{Dr. Dobb's Journal of Software Tools}, 2000.

\bibitem{Sucan2012TheOM}
I.~Sucan, M.~Moll, and L.~Kavraki, ``The open motion planning library,'' \emph{IEEE Robotics \& Automation Magazine}, vol.~19, pp. 72--82, 2012.

\bibitem{hart1968astar}
P.~E. Hart, N.~J. Nilsson, and B.~Raphael, ``A formal basis for the heuristic determination of minimum cost paths,'' \emph{IEEE Transactions on Systems Science and Cybernetics}, vol.~4, no.~2, pp. 100--107, 1968.

\bibitem{xin2022vision}
L.~Xin, Z.~Tang \emph{et~al.}, ``Vision-based autonomous landing for the uav: A review,'' \emph{Aerospace}, vol.~9, no.~11, p. 634, 2022.

\bibitem{saripalli2002vision}
S.~Saripalli, J.~F. Montgomery, and G.~S. Sukhatme, ``Vision-based autonomous landing of an unmanned aerial vehicle,'' in \emph{Proceedings 2002 IEEE international conference on robotics and automation (Cat. No. 02CH37292)}, vol.~3.\hskip 1em plus 0.5em minus 0.4em\relax IEEE, 2002, pp. 2799--2804.

\bibitem{tsai2006terminal}
A.~C. Tsai, P.~W. Gibbens, and R.~H. Stone, ``Terminal phase vision-based target recognition and 3d pose estimation for a tail-sitter, vertical takeoff and landing unmanned air vehicle,'' in \emph{Advances in Image and Video Technology: First Pacific Rim Symposium}.\hskip 1em plus 0.5em minus 0.4em\relax Springer, 2006, pp. 672--681.

\bibitem{lange2009vision}
S.~Lange, N.~Sunderhauf, and P.~Protzel, ``A vision based onboard approach for landing and position control of an autonomous multirotor uav in gps-denied environments,'' in \emph{2009 international conference on advanced robotics}.\hskip 1em plus 0.5em minus 0.4em\relax IEEE, 2009, pp. 1--6.

\bibitem{garrido2014automatic}
S.~Garrido-Jurado, R.~Mu{\~n}oz-Salinas, F.~J. Madrid-Cuevas, and M.~J. Mar{\'\i}n-Jim{\'e}nez, ``Automatic generation and detection of highly reliable fiducial markers under occlusion,'' \emph{Pattern Recognition}, vol.~47, no.~6, pp. 2280--2292, 2014.

\bibitem{verma2024comparative}
S.~Verma, A.~K. Sharma, and T.~Sandhan, ``Comparative analysis of fractal and aruco marker for navigation and landing of drones,'' in \emph{2024 15th International Conference on Computing Communication and Networking Technologies (ICCCNT)}.\hskip 1em plus 0.5em minus 0.4em\relax IEEE, 2024, pp. 1--7.

\bibitem{blachut2022automotive}
K.~Blachut, M.~Danilowicz, H.~Szolc, M.~Wasala, T.~Kryjak, and M.~Komorkiewicz, ``Automotive perception system evaluation with reference data from a uav’s camera using aruco markers and dcnn,'' \emph{Journal of Signal Processing Systems}, vol.~94, no.~7, pp. 675--692, 2022.

\bibitem{li2020aruco}
B.~Li, J.~Wu, X.~Tan, and B.~Wang, ``Aruco marker detection under occlusion using convolutional neural network,'' in \emph{2020 5th international conference on automation, control and robotics engineering (CACRE)}.\hskip 1em plus 0.5em minus 0.4em\relax IEEE, 2020, pp. 706--711.

\bibitem{zhao2023vision}
Z.~Zhao, J.~Lee, Z.~Li, C.~H. Park, and P.~Wei, ``Vision-based perception with safety awareness for uas autonomous landing,'' in \emph{AIAA SCITECH 2023 Forum}, 2023, p. 0126.

\bibitem{Zhou2020EGOPlannerAE}
X.~Zhou, Z.~Wang, C.~Xu, and F.~Gao, ``Ego-planner: An esdf-free gradient-based local planner for quadrotors,'' \emph{IEEE Robotics and Automation Letters}, vol.~6, pp. 478--485, 2020.

\bibitem{Zhu2021TPHYOLOv5IY}
X.~Zhu, S.~Lyu, X.~Wang, and Q.~Zhao, ``Tph-yolov5: Improved yolov5 based on transformer prediction head for object detection on drone-captured scenarios,'' \emph{2021 IEEE/CVF International Conference on Computer Vision Workshops (ICCVW)}, pp. 2778--2788, 2021.

\bibitem{shah2018airsim}
S.~Shah, D.~Dey, C.~Lovett, and A.~Kapoor, ``Airsim: High-fidelity visual and physical simulation for autonomous vehicles,'' in \emph{Field and Service Robotics: Results of the 11th International Conference}.\hskip 1em plus 0.5em minus 0.4em\relax Springer, 2018, pp. 621--635.

\bibitem{Hornung2013OctoMapAE}
A.~Hornung, K.~M. Wurm, M.~Bennewitz, C.~Stachniss, and W.~Burgard, ``Octomap: an efficient probabilistic 3d mapping framework based on octrees,'' \emph{Autonomous Robots}, vol.~34, pp. 189 -- 206, 2013.

\bibitem{unrealengine}
\BIBentryALTinterwordspacing
{Epic Games}, ``Unreal engine.'' [Online]. Available: \url{https://www.unrealengine.com}
\BIBentrySTDinterwordspacing

\bibitem{kalaitzakis2020experimental}
M.~Kalaitzakis, S.~Carroll, A.~Ambrosi, C.~Whitehead, and N.~Vitzilaios, ``Experimental comparison of fiducial markers for pose estimation,'' in \emph{2020 International Conference on Unmanned Aircraft Systems (ICUAS)}.\hskip 1em plus 0.5em minus 0.4em\relax IEEE, 2020, pp. 781--789.

\bibitem{Karaman2011SamplingbasedAF}
S.~Karaman and E.~Frazzoli, ``Sampling-based algorithms for optimal motion planning,'' \emph{The International Journal of Robotics Research}, vol.~30, pp. 846 -- 894, 2011.

\bibitem{jetsonnano}
NVIDIA, ``Jetson nano devkit,'' \url{https://www.nvidia.com/en-au/autonomous-machines/embedded-systems/jetson-nano/product-development/}, 2019.

\bibitem{tensorrt}
Nvidia, ``Tensorrt,'' \url{https://github.com/NVIDIA/TensorRT}, 2024.

\bibitem{F450}
H.~World, ``F450 drone kit,'' \url{https://www.hawks-work.com/products/f450-drone-kit-to-build-diy-450mm-wheelbase-4-axis-multi-rotor-drone-kit-d}, 2024.

\bibitem{pixhawk}
Ardupilot, ``pixhawl overview,'' \url{https://ardupilot.org/copter/docs/common-pixhawk-overview.html}, 2024.

\bibitem{Realsense_D435}
Intel, ``Realsense depth camera d435,'' \url{https://www.intelrealsense.com/depth-camera-d435/}, 2024.

\bibitem{neo3}
CUAV, ``Neo 3 gnss modules,'' \url{https://www.cuav.net/en/neo-3-en/}, 2024.

\bibitem{TFMini-Plus}
Benewake, ``Tfmini plus,'' \url{https://en.benewake.com/TFminiPlus/index.html}, 2024.

\bibitem{zheng2025neurostrata}
X.~Zheng, Z.~Li \emph{et~al.}, ``Neurostrata: Harnessing neurosymbolic paradigms for improved design, testability, and verifiability of autonomous cps,'' 2025.

\end{thebibliography}

\end{document}